\documentclass[12pt,twoside]{article}
\usepackage[
  margin=2.22cm
]{geometry}
\usepackage{graphicx} 
\usepackage{amsmath}
\usepackage[T1]{fontenc}
\usepackage[utf8]{inputenc}

\usepackage[misc]{ifsym}

\usepackage[hyphens]{url}

\usepackage{natbib}

\providecommand{\parencite}{\citep}
\providecommand{\textcite}{\citet}

\usepackage{hyperref}

\usepackage[shortcuts]{extdash}

\usepackage{enumitem}
\newlist{argument}{enumerate}{1}
\setlist[argument]{label=(\arabic*), labelindent=1.7em, leftmargin=*,
                   labelsep=0.5em, align=left,
                   topsep=0.45em, itemsep=0.32em, parsep=0pt, partopsep=0pt}
\newcommand{\arghead}[1]{\par\medskip\noindent\hspace*{1.7em}\textbf{#1}\par\nobreak}

\usepackage{float}

\usepackage{xcolor}

\usepackage[most]{tcolorbox}
\tcbset{
  abstractbox/.style={
    colback=gray!10,     
    colframe=gray!10,   
    boxrule=0.3pt,
    arc=1mm,
    left=6pt,
    right=6pt,
    top=4pt,
    bottom=4pt,
    boxsep=5pt
  },
  defbox/.style={
    colback=gray!5,
    colframe=gray!5,
    boxrule=0.3pt,
    arc=1mm,
    left=6pt,
    right=6pt,
    top=4pt,
    bottom=4pt,
    boxsep=5pt
  }
}

\definecolor{refcolor}{rgb}{0.1, 0.1, 0.4}

\hypersetup{
    colorlinks=true,        
    linkcolor=refcolor,     
    citecolor=refcolor,     
    urlcolor=refcolor       
}

\usepackage{crimson}

\usepackage[activate={true,nocompatibility},final,auto=true,tracking=true,kerning=true,expansion=true,spacing=true,factor=1000,stretch=20,shrink=20,letterspace=0]{microtype}

\usepackage[font=small,labelfont=bf]{caption}

\usepackage{fancyhdr}
\usepackage{lastpage}

\newcommand{\doctitle}{A Polyphonic Conception of AI Understanding}

\fancypagestyle{plain}{%
  \fancyhf{} 
  \fancyhead[L]{September 2026}
  \fancyhead[R]{\textit{Manuscript}}
}

\usepackage{authblk} 

\title{\textbf{A Polyphonic Conception of AI Understanding}}
\author[1]{Matthieu Queloz}
\author[2,3,4]{Pierre Beckmann}
\affil[1]{University of Bern, Department of Philosophy}
\affil[2]{École Polytechnique Fédérale de Lausanne (EPFL)}
\affil[3]{Idiap Research Institute}
\affil[4]{Machine Alignment, Transparency, and Security (MATS)}

\affil[ ]{~\vspace{-0.2cm}}
\affil[ ]{\textit{Both authors contributed equally.}}

\affil[ ]{~\vspace{-0.2cm}}

\affil[ ]{\textcolor{gray}{\small\Letter} \hspace{0.05cm} \href{mailto:matthieu.queloz@unibe.ch}{\texttt{matthieu.queloz@unibe.ch}}}

\makeatletter
\renewcommand{\maketitle}{\bgroup\setlength{\parindent}{0pt}
\thispagestyle{plain}
\begin{flushleft}
\vspace*{0.7cm}
  {\fontsize{17}{17}\selectfont \@title}
  \vspace{0.5cm}

  \@author

\end{flushleft}\egroup
}
\makeatother

\usepackage{titlesec}

\titleformat{\section}
  {\fontsize{17}{15}\selectfont\bfseries}
  {\thesection}{1em}{}

\begin{document}

\vspace{1cm}

\maketitle

\vspace{0.2cm}

\begin{tcolorbox}[abstractbox]
\noindent
\textbf{Abstract:}
When a doctor, a judge, or an engineer must decide whether to trust an AI model's output, they cannot avoid asking what the model \textit{understands}. Purely mathematical or statistical descriptions struggle to distinguish trustworthy from untrustworthy outputs without reintroducing the question of AI understanding in all but name. Yet the question is ill-framed as it stands, because the inherited concept operates within a \textit{monophonic} paradigm: the idea that a cognitive system’s understanding of something must be localised to a single mechanism underpinning all the capacities conferred by such understanding. Drawing on a wide range of mechanistic evidence, we show that LLMs are pervasively \textit{polyphonic}: outputs emerge from coalitions of parallel mechanisms of uneven reliability, which variously complement, duplicate, or drown out one another, with several coalitions sufficing for a task without any one being indispensable. Polyphony not only complicates attributions of understanding, but renders monophonic inference patterns hazardous. In response, we develop a conception of understanding fit for polyphonic AI. It centres on sound circuitry that is reliably and correctly recruited and in control of outputs. Attributions of understanding thereby become tractable claims about internal organisation, and can do the work of guiding trust in AI.

\vspace{0.3cm}
\noindent

\end{tcolorbox}

\vspace{0.0cm}

\section{The Jury Room Analogy} \label{section:1}

Imagine a jury room in which deliberation is conducted in relays over several days, with a fresh panel of jurors coming in every morning. What persists are the notes that each  panel leaves on the large communal table. Incoming jurors read this material selectively and with varying degrees of competence. Some happen to have occupational expertise they can bring to bear; others reason more superficially, judging witnesses by their manner, or counting how often a name recurs in the evidence. Over time, accurate assessments and arguments accumulate on the table alongside mistakes and misreadings. There is no presiding juror to orchestrate the proceedings. Nor does the verdict depend on a particular set of jurors: had one cluster of them stayed home, others would have taken up their share of the work. Yet despite this multitude of contributing voices, the relay jury is, in the end, forced to deliver one verdict with one voice.

Does the jury \textit{understand} the case? After all, not a single juror had a comprehensive grasp of the entire case. The verdict was not the direct expression of a single locus of understanding. It emerged from the complex interplay of hundreds of voices of uneven quality. The question this raises is not the familiar one from social epistemology -- what to say when many minds coalesce into one collective agent \parencite{list2011group, bird2010social} -- but rather what to say when what \textit{presents} as a unified individual mind disaggregates, upon closer inspection, into many voices or sub-personal mechanisms.

The relay jury offers a useful analogy for thinking about understanding in large language models (LLMs). LLMs process information through successions of layers that each deploy their own collection of parallel mechanisms. These mechanisms read from the communal table -- the model's \textit{residual stream}. Each mechanism looks for different things and writes its own conclusions back into the stream for later layers to read. Interpretability research has shown these mechanisms to be uneven in quality and highly selective in what they attend to \citep[see, e.g.,][]{elhage_mathematical_2021, lindsey_biology_2025}. Like the jury room, the model must return a single verdict, which conceals the multitude of voices it contains. We call this condition \textit{polyphony}, in an echo of the philosopher Mikhail Bakhtin's account of Dostoevsky's novels, which hold many independent voices and perspectives without subordinating them to a single authorial vision, leaving the whole to emerge from their interplay.\footnote{See also \cite{shaw2026polyphonic}, whose formally oriented account of ``polyphonic intelligence'' provides independent evidence of the fruitfulness of the metaphor. Whereas Shaw develops a theoretical proposal for non-dominating integration among plural inferential processes, we draw on mechanistic findings about existing LLMs to explore the implications of polyphony for attributions of understanding.}

We contend that the polyphonic character of LLMs productively complicates the recent debate over whether such systems can be said to understand. The positions in that debate fall into three camps. One camp -- call them the ``anthropomorphisers'' -- unabashedly applies cognitive terms like ``understanding'' to LLMs.\footnote{See \cite{piantadosi2022meaning, bubeck2023sparks, sogaard2023grounding, mandelkern2024refer, brooks2024video, cappelen2025whole}.} A second camp -- the ``deflationists'' -- admonishes against the use of cognitive terms in this connection, maintaining that LLMs are best conceptualised in statistical or mathematical terms -- we should speak of ``matrix multiplications'' producing outputs from ``superficial correlations'', ``distributional statistics'', or ``pattern-matching''.\footnote{See \cite{chomsky2023false, shanahantalking1, titus2024chatgpt, yiu2023imitation, bender2021dangers, marcus2018deep, floridi2023ai, bishop2021artificial, bender2025aicon}.}

Though it is more often practised than defended in print, there is a third position, which dismisses the entire debate as a distraction: as long as the models produce the outputs we want, these ``indifferentists'' maintain, we should not worry too much about how they got there.\footnote{See \cite{krishnan2020against, carlini2025machines}.} Whether AI models ``understand'' is a philosophical question in the pejorative sense of the term: an idle speculation.

We argue that the question of AI understanding is anything but idle -- it \textit{needs} to be raised even by the most hard-nosed practitioners, because it \textit{guides the allocation of trust}. Yet making headway with it requires breaking out of what we call the \textit{monophonic paradigm}: the perennially tempting idea that a cognitive system's understanding of a subject matter must be localised to one substrate, a \textit{locus of understanding}, which simultaneously underpins all the capacities that this understanding confers and reliably finds expression in them.

This monophonic paradigm works well enough in its home territory of human cognition. Thinking of many capacities as tracing back to one seat of understanding helpfully licenses the largely reliable inference from an individual's display of competence in one respect -- in \textit{explaining} something, for instance -- to the presumption that this individual will prove competent in other respects as well, i.e.\ in controlling, diagnosing, predicting, or reasoning counterfactually about that thing. Conversely, failure at any of these is reasonably taken to count against the presence of understanding more broadly (e.g.\ ``if you can't explain it, you don't really understand it'').

When transposed to LLMs, however, the monophonic paradigm leads us astray, because the inferences it underwrites no longer reliably hold. Conversing with an LLM may \textit{feel} like interacting with a unified mind that speaks with a single voice, but mechanistic inspection reveals that todays's LLMs, at least, are pervasively polyphonic systems. As Jacob Andreas puts it in a congenial essay, LLMs are best conceptualised as ``collections of competing mechanisms -- a knowledge retrieval circuit voting against a copying circuit voting against an n-gram model'' (\citeyear{andreas2024world}). This polyphonic character is visible already at the level of individual representations. What we would ordinarily treat as a single concept may be represented in several partly overlapping ways; conversely, the same part of the model may contribute to several different representations or tasks. But polyphony becomes even more striking when individual represenations are chained together to form complex procedures: multiple pathways may have to combine to produce an answer, while the same task may also be discharged by alternative, sometimes overlapping combinations of pathways. Their contributions can reinforce, inhibit, replace, or compensate for one another. Every word an LLM produces reverberates with many voices --- and they are not singing in unison.

We argue that a conception of understanding capable of making sense of LLMs must be a correspondingly polyphonic one. We prefer ``polyphonic'' to ``distributed'' \parencite{hutchins1995cognition}, because distributed cognition involves spreading a task outward across many agents and artifacts, whereas polyphony involves a spreading inwards, a subdivision into multiple mechanisms that look, from the outside, like a single agent. Nor is the relevant form of understanding best described as ``fragmented'' or ``fractured'', as in Freeborn's (\citeyear{freeborn2026fractured}) account of AI understanding. This evokes a lost unity -- yet polyphony is not necessarily a defect. The relay-jury structure is its own kind of unity. It constitutes a distinct form of understanding. As Andreas (\citeyear{andreas2024world}) argues, it can be a source of efficiency -- LLMs would not get much done in a single forward pass through the network if they could not occasionaly rely on quick heuristics. And as we shall show, polyphony can also be a source of robustness and accuracy.

A significant consequence of polyphony, however, is that an outwardly competent performance may be underpinned by a multitude of criss-crossing mechanisms, none of which rise to a level we would ordinarily want to dignify with the term ``understanding''. Conversely, a disappointing performance may conceal truly ingenious and sound procedures whose contributions get drowned out by competing signals. The monophonic paradigm, which encourages us to treat competent behaviour as testimony to a single underlying locus of understanding, licenses a number of inferences that become hazardous in the face of such polyphonic systems. As \textcite{milliere2026anthropocentric} argue, this is a form of \textit{anthropocentric bias}: we transfer to AI models a conception of understanding that is keyed to human cognition, and are led astray by insufficient appreciation of how profoundly these systems can differ from us.

It will be objected that human cognition is itself not monophonic: the brain does many things at once, and while philosophy has tended to emphasise the \textit{unity} of consciousness \parencite{bayne2010unity}, theories of cognition have found uses for pictures of a divided mind, from Minsky's (\citeyear{minsky1986society}) society of mind through Dennett's (\citeyear{dennett1991consciousness}) multiple drafts to Kahneman's (\citeyear{kahneman2011thinking}) fast and slow systems.\footnote{See \cite{evans2013dual} for an overview of dual-process theories of higher cognition.} But the point is that the monophonic paradigm works in practice for human cognition. Whether through internal coordination, external scaffolding, or a mixture of both, we have learned to fashion human individuals into suitable units for attributions of understanding -- beings in whom the relevant capacities keep company reliably enough to treat them as ramifications of a single seat of understanding. This fashioning of a polyphonic interior into a stable unit has not (yet) happened for LLMs to the same extent. This is one reason why the inherited concept of understanding does not quite fit them.\footnote{Though we focus on understanding, polyphony also complicates other cognitive attributions at the level of the model as a whole, including attributions of belief, intention, and knowledge.}

Here, we propose to remedy this lack of fit by re-engineering the concept of understanding for LLMs in light of the mechanistic evidence of their polyphony. We thereby distance ourselves from all three camps in the AI understanding debate. Against the indifferentists, we argue that the question whether AI understands is anything but idle: it does need to be asked, and a great deal turns on the answer; while it is true that one's answer depends on one's conception of understanding, the conclusion to draw is that the debate needs to ascend to the metaconceptual level and face the question of \textit{how best to conceptualise} understanding in the context of AI systems. Against the deflationists, we maintain that the cognitive register, and some conception of understanding in particular, is \textit{indispensable}: as our \textit{reintroduction argument} establishes, mathematical and statistical vocabulary is too indiscriminate to separate trustworthy from untrustworthy outputs, and any vocabulary rich enough to do so ends up reintroducing the concept of understanding in all but name. Against the anthropomorphisers, we note that our existing concept of understanding is \textit{inadequate}, because it carries monophonic assumptions that do not transfer to LLMs. We thus need to \textit{adapt} the way we conceptualise understanding. Drawing on a range of recent findings in mechanistic interpretability research, we show how a suitably adjusted conception of understanding can gain a foothold in the actual workings of LLMs and do much-needed work in our interactions with AI.

We proceed as follows: \S\ref{section:2} shows why we \textit{need} a coherent way to think about understanding in LLMs if we are to know when to trust their outputs. \S\ref{section:3} lays out what it would mean for mechanistic organisation to vindicate such trust. As \S\ref{section:4} then brings out, however, LLM understanding is pervasively polyphonic, which complicates attributions of understanding in several respects. \S\ref{section:5} then addresses how these complications play out at the scale of frontier models and synthesises the main ways in which we can determine whether to trust AI outputs. \S\ref{section:6} concludes by considering the prospects for orchestrating polyphony.

\section{The Reintroduction Argument} \label{section:2}

The question of how to conceptualise AI cognition is not merely a verbal dispute. In a growing range of situations, people must decide whether and how far to trust AI outputs. We contend that they cannot do so without forming a view of what the model has and has not understood.

Consider a doctor consulting an AI model about a perplexing case. Suppose the model proposes an intriguing diagnosis. The doctor must then decide how far to trust it. For this, she needs to know whether the model has \textit{understood} the disease: whether it has grasped how the disease's causes, symptoms, and treatments hang together, or whether it is merely parroting medical language. The concept of understanding does indispensable work here: it guides the allocation of epistemic trust.

The doctor's predicament is far from unique. A judge weighing AI-generated sentencing recommendations must discriminate between recommendations grounded in a grasp of how precedents bear on the present case and ones that just pattern-match based on surface features; a professor deciding whether to direct students to an AI tutor must determine whether the model's command of the material warrants an endorsement. These practitioners cannot afford to shirk these discriminative tasks by dismissing the understanding question as idle philosophising. They have a pressing \textit{conceptual need} -- an instrumental need for a conceptual resource capable of doing the work traditionally performed by the concept of understanding. And if that work essentially involves differentiating \textit{between} AI systems and their outputs, this also means that the question cannot remain whether the inherited concept of understanding applies to AI systems \textit{as a class}. The question must be how our inherited concept of understanding needs to be adapted to perform discriminative labour \textit{within} that class.

This is why indifferentism is unsustainable as a response to the understanding question. The question is anything but idle: it \textit{demands} an answer because a lot turns on it. And the point is not that \textit{theorists} need to address it; it is \textit{practitioners} -- doctors, judges, and teachers -- that need to address it. The question is ineluctable \textit{in practice}.

Now deflationism, on the other hand, requires a more complex response, because deflationists may grant the ineluctability of the question, but insist that with LLMs, it uniformly receives a negative answer: LLMs can \textit{never} properly be said to ``understand.'' Indeed, one can exhaustively describe what they do without employing any cognitive vocabulary. A transformer converts its input into numerical ``embeddings'', passes them through layers of linear projections and non-linear transformations, and returns a probability distribution over possible next tokens. An LLM is thus a complicated function from input text to a probability distribution over continuations.

Such mathematical descriptions are perfectly correct, as far as they go. The problem is that they do not go far enough: they are too undiscriminating. They apply equally whether the model gives an accurate diagnosis or a ludicrous one -- whether it solves a problem using a robust and general procedure or using a fragile heuristic. Mathematical descriptions apply across the cases we need to distinguish, without supplying the nuances required to tell those cases apart. Even a complete mathematical specification of a forward pass through the model would not, by itself, tell us whether the model is matching superficial linguistic patterns or deploying anything like joint-carving concepts organised to mirror the structure of a domain. Yet that is the question that matters. And any serious attempt to address it is bound to reintroduce the concept of understanding in all but name.

Consider the doctor again. Even if she takes the deflationist's advice and tries to think about her predicament entirely in non-cognitive terms -- purely in terms of pattern-matching, for instance -- she must decide whether to trust the model's output. And this forces a discrimination between different \textit{kinds} of pattern-matching. Not all pattern-matching is equal: some forms merit deference; others do not. But what \textit{is} the kind of pattern-matching that merits deference? It is, one might think, pattern-matching that \textit{tracks the underlying structure of the domain} -- notably, by being sensitive to the causal dependencies between aetiological factors, pathogenetic mechanisms, and clinical manifestations rather than just to superficial correlations between them.

The deflationist can of course speak of ``structure-tracking pattern-matching'' and link that to epistemic trustworthiness; but the \textit{distinction} she thereby draws, and the inferential implication she ties it up with, already begins to reproduce the inferential profile of the understanding/parroting distinction. The concept of understanding is being reintroduced through the back door.

The heart of what we call \textit{the reintroduction argument}, then, is that any attempt to draw the distinction between reliable and unreliable AI outputs in deflationary vocabulary will end up reforging just the inferential connections that are characteristic of the vocabulary one seeks to avoid, thereby effectively reintroducing the concept of understanding, or at least a recognisable descendent of it, in all but name.

The fundamental reason for this is that non-cognitive vocabulary is \textit{expressively inadequate} to the task of discriminating between the finer shades of competence and reliability that we need to distinguish in dealing with systems as sophisticated as contemporary AI models.

To see this, consider what happens if our doctor recognises the need for finer conceptual discrimination, but remains resolved to avoid cognitive vocabulary. She could look to \textit{track records of success}. These are readily available as benchmark statistics: a 95\% score on a medical-diagnosis benchmark (such as CPC-Bench, MSDiagnosis, or the Sequential Diagnosis Benchmark) would seem to offer some grounds for trust.

Yet success on benchmarks remains compatible with extensive reliance on shortcuts and superficial cues. A track record of success is too coarse an indicator to distinguish between a trustworthy and an untrustworthy model with 95\% score. Even if that track record extends across a diversity of benchmarks and datasets, a gap remains between good performance on tests and real-world reliability on a novel case.

Our doctor needs to know whether the model can be trusted in \textit{this particular real-world case}. She may therefore want to consider the model's real-world track record across cases that are relevantly similar to the case at hand. Yet even supposing real-world data to be available, this begs the question of which cases to \textit{count} as relevantly similar. It was because the case was unusual and unobvious that she consulted the model to begin with; and \textit{which} cases are relevantly similar to this one depends on the causal structure of the disease at hand -- which is precisely what she is trying to find out in asking the model, and thus cannot help her in deciding whether to trust the model's answer.

What these attempts to evaluate a model using its performance history are fundamentally missing is sensitivity to the \textit{internal basis} of the model's performance. Only something that \textit{persists} across cases can underwrite trust in the model's competence on a novel case. One thus needs to look at what goes on inside the model -- at the output-producing \textit{process}.

Yet the moment one tries to identify trustworthiness-conferring properties of that process, one starts to recapitulate the literature's attempts to articulate what understanding is. Suppose the doctor asks whether the model is sensitive to the domain's \textit{difference-makers} -- whether its outputs covary with what an intervention on the disease would change. That is the mark of understanding on causal accounts, which characterise understanding as a grasp of what-would-happen-if relationships and of the factors on which an outcome depends \parencite{pearl2018book, woodward2003making, strevens2008depth}. Suppose she asks whether the model's success is \textit{robust} -- surviving structure-preserving variations and extending to relevantly novel cases. That is understanding as cognitive control: a cluster of abilities expressing mastery of the relationships between \textit{p} and the reasons why \textit{p} \parencite{hills2016understanding, de_regt2017scientific}. Suppose she asks whether the model's outputs on related questions \textit{hang together} -- whether its diagnosis coheres with what it says about the disease's course, complications, and treatment. That is understanding as grasping connections \parencite{Wittgenstein1953-WITPI-4, riggs2003balancing, grimm2011understanding, elgin2017true, kvanvig2018knowledge}. Or suppose she asks whether the model has \textit{compressed} its cases into an underlying rule. That is understanding as unification \parencite{friedman1974explanation, kitcher1989unification, schurz1994outline, wilkenfeld2018compression, freeborn2026fractured}.

The list could be extended, but the reintroduction argument concerns the pattern it instantiates. The literature on understanding maps out the inferential terrain characteristic of the concept. And while accounts diverge over which region of the terrain they regard as central, the point is that any discriminative resource that rises to the doctor’s challenge must occupy some region of that terrain. It may do so without employing the word ``understanding'', but it will at least partly reenact the inferential role played by that word. And whether or not we take inferential role to \textit{individuate} concepts, these inferential connections are what the debate over AI understanding is about; a deflationism that reintroduces those connections has effectively conceded the debate.

The cognitive register is thus hardly an indulgence; it is the register in which the relevant facts come into view.

\section{Circuitry that Licenses Trust} \label{section:3}

What anthropomorphisers neglect, however, is that a concept can be indispensable and still unfit as it stands. The inherited concept earns its extension to AI through the important inferential connections it draws. But it also carries inferential connections that reflect its human origins. Re-engineering the concept for AI is a matter of sorting out which connections need \textit{preserving} because they continue to do indispensable work, which need \textit{forging} so that they anchor the concept in the distinctive make-up of models, and which need \textit{severing} because they do not transfer to AI. Millière and Rathkopf anticipate that investigating LLM capacities will yield ``a novel ontology of cognitive kinds, optimized for explaining the distinctive strengths and weaknesses of machine intelligence rather than human intelligence'' (\citeyear{milliere2026anthropocentric}, 386). What follows is an attempt to supply one entry in that novel ontology.

If attributions of understanding to AI are to be more than behaviourist courtesy, they must gain a foothold in what models actually contain. Following \textcite{beckmann2026mechanistic}, we take the mechanistic basis for LLM understanding to comprise three kinds of structure: \textit{features}, \textit{connections} between features, and \textit{circuits} encoding more or less general procedures.

LLM layers read from and write to a shared internal state -- the residual stream. That state is a long list of numbers that can be pictured as a latent \textit{space}, in which each number configuration picks out a point. As the numbers change, that point shifts in a corresponding direction.

It appears that LLMs encode concepts by letting each direction in the space stand for something: they treat the directions as representing the degree to which something exhibits a certain feature. Researchers identified a direction that stands for the presence of the Golden Gate Bridge across languages, paraphrases, and imagery, for example \parencite{templeton_scaling_2024}. These internal representations have become known as ``features.'' \textcite{yetman2026} and \textcite{williams2026} point out that these features must be causally efficacious in the right way to count as representations. Their causal efficacy can be demonstrated through \textit{steering}: if one moves the point picked out by the list of numbers further along the direction representing the Golden Gate bridge, the LLM starts fixating on the bridge.

\begin{figure}[h]
    \centering
    \includegraphics[width=\linewidth]{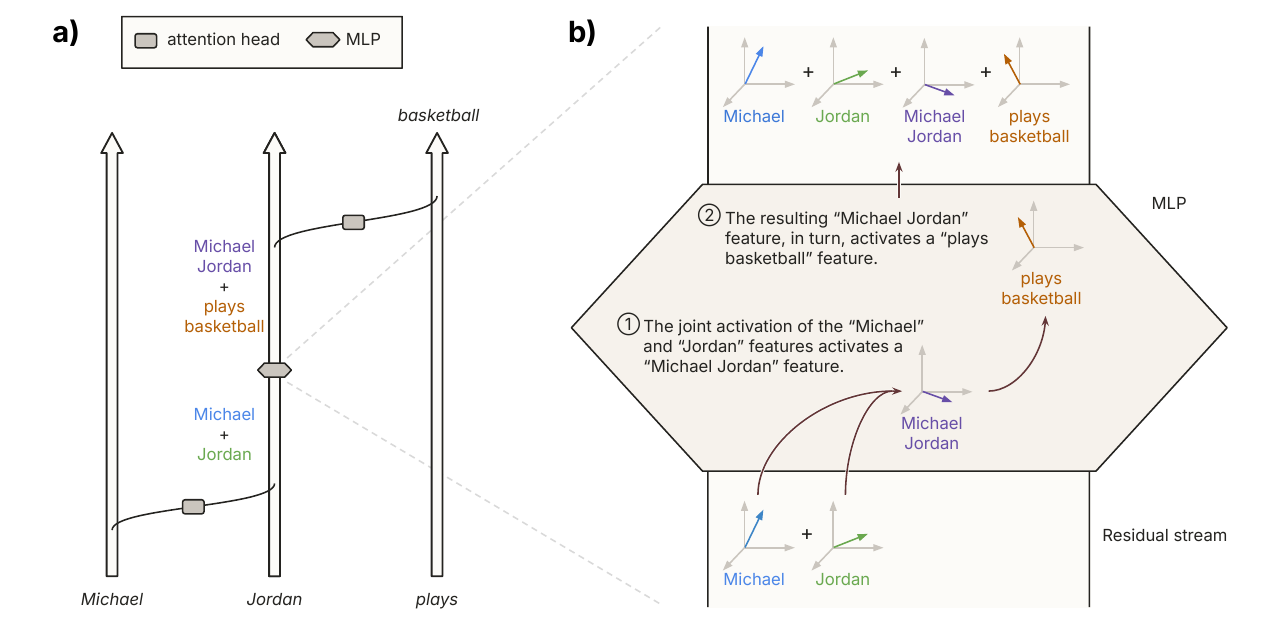}
    \caption{Connections between features ensure that the model completes the sequence ``Michael Jordan plays'' with ``basketball''. The main role of attention heads is to carry forward features between residual streams (a). MLPs mainly combine features and recall features via activated features (b). Features are activated via directions in the residual stream's latent space.}
    \label{fig:connections}
\end{figure}

To learn facts about the world, an LLM needs to learn to \textit{connect} its features in the right way. Connections between features take the form of learned dependencies by which activating one feature drives up (or down) the activation of another. This enables a model to represent the fact that there is a connection between Michael Jordan and basketball, for example, which allows it to retrieve the fact that Michael Jordan is a basketball player (Figure~\ref{fig:connections}).

A \textit{circuit}, finally, is a set of model components, typically scattered across several layers, that are wired together to implement some less content-specific \textit{procedure} instead of relying on memorised facts. A well-understood example is a circuit formed to compute modular addition \parencite{nanda_modular_addition_2023, chughtai2023toy, li2025modular, beckmann2026mechanistic}.

We propose to use the term \textit{sound circuitry} to refer to a constellation of features, connections, and circuits that licenses trust in an AI model -- relative to a certain type of task in a certain domain and across a certain range of cases. Such a constellation is \textit{sound} to a degree insofar as, under an independently warranted interpretation of its causally efficacious states, those states bear a non-accidental, approximately structure-preserving relation to the elements and dependencies of the domain that matter to the task, and its transitions exploit that relation so that, were the constellation recruited and allowed to govern the computation, it would meet the correctness standard with corresponding reliability across the specified scope. \textit{Soundness} is thus a graded term of art rather than the logician's all-or-nothing notion.

The conception of AI understanding we want to develop centres on sound circuitry as what gives the notion of understanding a grip on the internal organisation of LLMs. Let us work through two case studies to substantiate this claim.

\subsection{Connecting Features for Medical Diagnosis} \label{subsection:3.1}

\textcite{lindsey_biology_2025} examined what goes on under the hood when an LLM performs a differential diagnosis. They gave Claude 3.5 Haiku the prompt: ``A 32-year-old female at 30 weeks gestation presents with severe right upper quadrant pain, mild headache, and nausea. BP is 162/98 mmHg, and labs show mildly elevated liver enzymes. If we can only ask about one other symptom, we should ask whether she's experiencing\dots'' -- and Claude provided ``visual disturbances'' as the likeliest completion, followed by ``proteinuria'' (too much protein in urine).

To a doctor, these are the right continuations. The question is how Claude gets there. Does it rely on co-occurrence patterns between words in the prompt and those continuations? Or does it do what a doctor does, i.e.\ recognise the symptoms as typical of \textit{preeclampsia} and then reason that hitherto unmentioned symptoms of preeclampsia include visual disturbances and proteinuria? The direct route requires no understanding of what is being diagnosed; the indirect route requires a grasp of the connections between the symptoms presented, the disease, and its further symptoms.

To determine which route Claude uses, one needs to know not only which features are active in the model, but what the connections between these features are, and which features activate which. There is a technique for exposing this internal wiring. It involves producing a so-called \textit{attribution graph} \parencite{ameisen_circuit_2025, lindsey_biology_2025}.

One difficulty in producing such a graph is that almost every neuron fires for a miscellany of things -- it is \textit{polysemantic}. This is because the network needs to represent far more concepts than it has neurons by packing them in \textit{superposition} \parencite{olah2020zoom, elhage_toy_2022}. To nonetheless home in on the features that are key to generating an output, one trains a second network -- a \textit{transcoder} -- to reproduce the first network's computation using replacement units engineered so that only a small fraction are active at any moment and each tends to be interpretable as encoding one concept. This gives us a sparse representation of the features that are active in this one inference.

But this does not yet tell us how these features \textit{hang together}, i.e.\ which features activate which. Because the transcoder's features enter the computation additively, one can calculate how much each feature \textit{directly contributes} to activating later features. By representing each feature as a node and each contribution as a line or ``edge'', one can then generate an attribution graph showing which features activated which on the path from prompt to output.

Initially, this graph depicts millions of edges. It needs pruning to a subgraph that preserves most of the computation. This gives us a sparse representation of the \textit{connections between features} involved in the computation.\footnote{The graph reconstructs the model only approximately, however. What it cannot capture gets bundled into uninterpretable ``error'' terms.}

\begin{figure}[h]
    \centering
    \includegraphics[width=\linewidth]{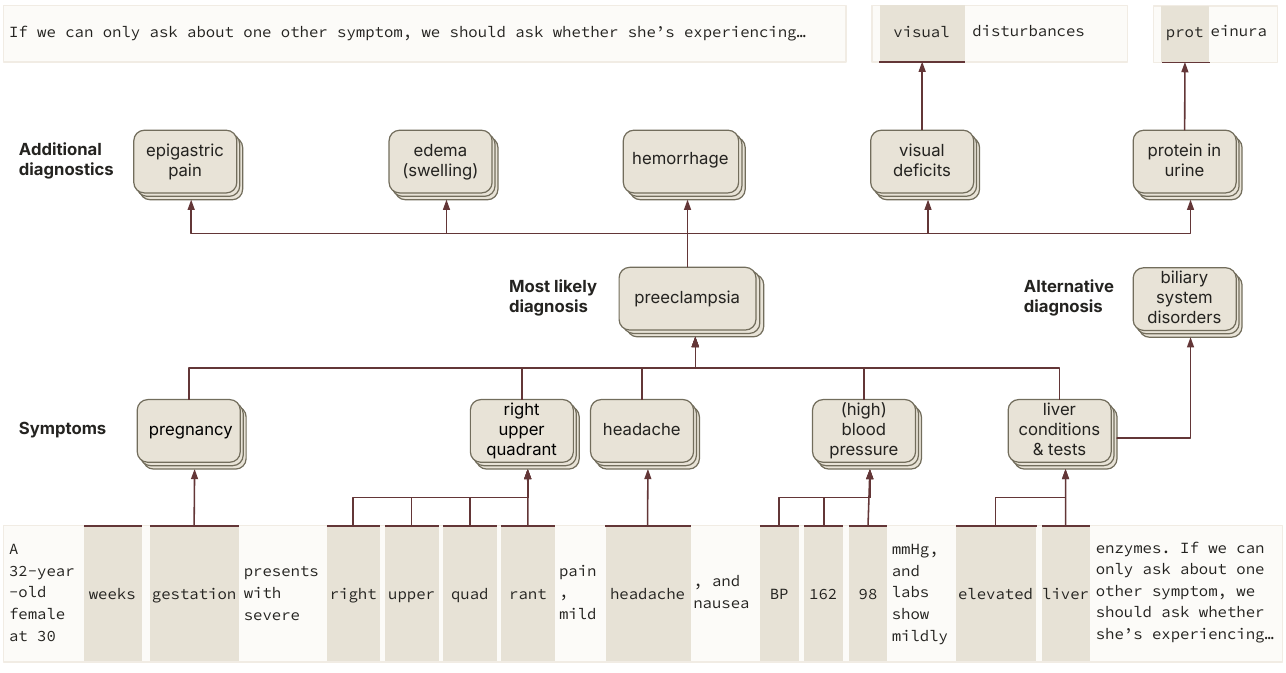}
    \caption{Attribution graph for Claude 3.5 Haiku \parencite{lindsey_biology_2025}. Patient-status features activate a \textit{preeclampsia} feature alongside competing hypotheses (\textit{biliary system disorders}); the \textit{preeclampsia} feature in turn activates features for confirmatory symptoms, which inform the model's leading completions: ``visual disturbances'' and the runner-up, ``proteinuria''.}
    \label{fig:attribution}
\end{figure}

An attribution graph can reveal whether the model took the superficial or the understanding-like route. If the model merely parroted medical text based on superficial statistics, the edges should run straight from the features for the prompt's words to the feature for ``visual disturbances''. On the understanding-like route, the edges should pass through diagnostic features such as the \textit{preeclampsia} feature. It turns out that Claude indeed takes that latter route (Figure~\ref{fig:attribution}).

To test whether the \textit{preeclampsia} feature is causally efficacious in producing the output rather than merely correlated with it, the experimenters inhibited that feature. The model's leading recommendation then flipped to asking about ``decreased appetite'' -- a sign of biliary system disorders, the model's alternative diagnosis. Not only has the model learned to wire up the connections between features to mirror the actual connections between preeclampsia and its symptoms; the model also brings features for rival hypotheses into play, much as a differential diagnosis would.

This is merely a look under the hood of a single inference, of course. Its ability to ground epistemic trust is correspondingly limited. It can warrant \textit{local} trust, for this type of diagnosis in that type of case, by showing that the answer issues from the right kind of connections between features. But these connections are content-specific. To warrant broader trust, a different kind of internal structure is needed.

\subsection{General Addition Circuits} \label{subsection:3.2}

What could warrant broader trust is a more general, more content-independent circuit. One of the first clear examples of this turned up in a small neural network trained to perform modular addition, i.e.\ addition with a ceiling (the \textit{modulus}) after which one starts over. Trained on a large table of examples, the network seemed to just memorise those. But when a researcher went on vacation and left the model training, they found on their return that accuracy on unseen examples had unexpectedly shot up after all. The model had also become internally simpler, replacing its memorised examples with a compact computational circuit \parencite{power2022grokking}.

Reverse-engineering this circuit, \textcite{nanda_modular_addition_2023} found that the model represents each number as an angle on a circle and adds these angles, so that the reset at the modulus happens automatically when coming full circle. This neat trick requires converting the landing point back into a number, however, which involves locating the peak of a relatively flat curve. To find this peak reliably, the model performs addition on multiple circles rotating at different speeds. Faster rotation sharpens the peak, yet also creates several of them. The model resolves that ambiguity by exploiting an effect analogous to what the physics of waves calls \textit{constructive interference}: it combines the curves in such a way as to suppress false peaks while amplifying the one on which all curves agree.

One might suspect such clever tricks to emerge only when training a small network on nothing but modular addition. In fact, however, the trick of representing numbers on several circles at once was found in larger, generalist models as well -- and even when computing \textit{regular} addition \parencite{nikankin_arithmetic_2025, kantamneni_language_2025}.

\begin{figure}[h]
    \centering
    \includegraphics[width=\linewidth]{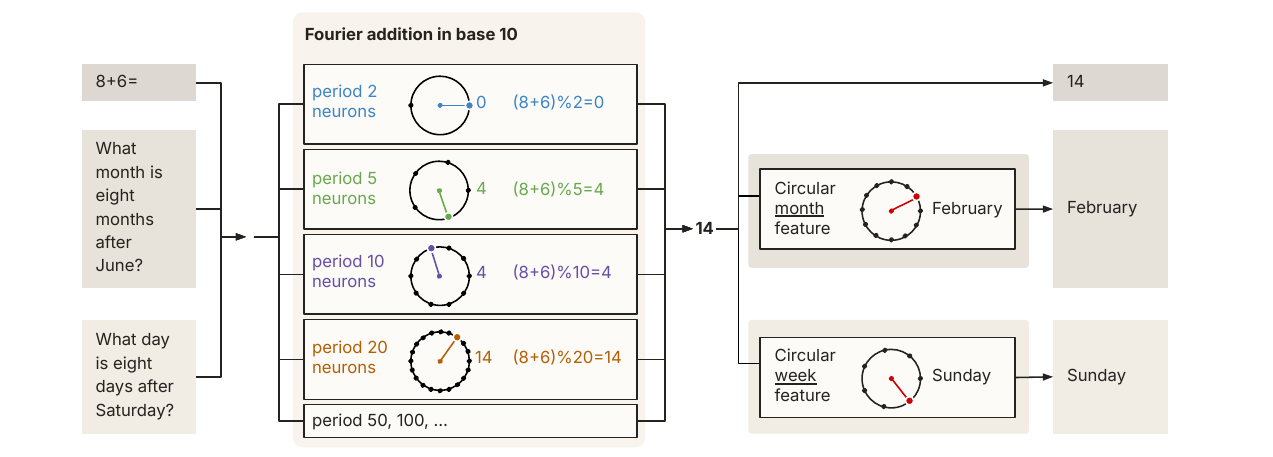}
    \caption{Instead of using separate calculators for each cyclic domain such as months, weekdays, or 24-hour times, Llama-3.1-8B reuses a single base-10 addition mechanism across addition-like tasks \parencite{feucht2026calculator}. This mechanism consists of several circles with periods of 2, 5, 10, 20, 50, and 100.}
    \label{fig:calculator}
\end{figure}

Llama-3.1-8B offers a particularly striking example of a general circuit along these lines. It was shown to reuse the same compact circuitry across addition-like tasks (Figure~\ref{fig:calculator}) -- not just regular addition, but also ``What month is six months after August?'' or ``What day is five days after Wednesday?'' \parencite{feucht2026calculator}. This is surprising, since models are known to represent cyclic concepts such as months, weekdays, and 24-hour times using dedicated circular representations with 12, 7, or 24 positions \parencite{zhou_pre-trained_2024, kantamneni_language_2025, feucht2026calculator}. Researchers expected the model to exploit these dedicated representations also when \textit{adding} months, weekdays, or 24-hour times. Computing ``four months after October'' on a circle with a period of 12 -- that is, twelve positions -- would, after all, make the \textit{modulo} 12 operation automatic.

Yet this is not what \textcite{feucht2026calculator} found. Llama-3.1-8B performs all kinds of additions -- whether involving plain arithmetic or months, weekdays, or hours -- in the same place: a circuit of just 28 neurons inside the MLP at layer 18. Cyclic concepts like months get translated into ordinary numbers and sent through a base-10 ``calculator''. Thus, ``What month is eight months after June?'' becomes $8+6$. The calculator uses circles for this, but not those with periods of 12, 7, and 24 that would be natural for months, weekdays, and hours. Instead, it uses circles with periods of 2, 5, 10, 20, 50, and 100. This decimal machinery can handle any kind of addition. But it, too, uses multiple circles in parallel to resolve ambiguities: on the period-5 circle, 14, the sum of $8+6$, is indistinguishable from 4, 9, and 19; it takes the period-20 circle to pin it down. In later layers, a \textit{modulo} 12 operation then maps the result -- 14 -- back to months, yielding the correct answer: February.

The discovery of such a circuit implementing a truly general and reliable algorithm supports trusting the model beyond this one case: insofar as that circuit continues to be recruited and to control outputs, Llama-3.1-8B can be expected to handle unseen examples across the various guises that addition takes. This is illustrative of the kind of broader trust that a general and reliable circuit can underwrite.

These examples of sound circuitry help substantiate the idea that attributions of understanding to AI models are, at bottom, claims about internal organisation. To say that a model \textit{understands} is to claim that, \textit{were} we to reverse-engineer what features, connections, and circuits are causally responsible for its performance, our findings would vindicate trust. The diagnosis and addition cases show that mechanistic interpretability can in principle uncover such organisation.

Yet the convenient singulars in which we described these findings -- \textit{the} preeclampsia feature, \textit{the} addition circuit -- conceal something important. Talk of ``the'' \textit{preeclampsia} feature is itself shorthand, used by researchers to group several internal representations that encode subtly different contents, but play roughly the same role in the computation -- a complication registered by the stacked boxes in Fig.~\ref{fig:attribution}. The general addition circuit likewise subdivides into multiple parallel computations that are ambiguous in isolation, but unambiguously point to an answer when considered together. In both cases, the singular label suggests a single, self-contained mechanism where we actually find an interacting plurality. This prefigures the argument of the next section: that LLMs are pervasively polyphonic.

\section{Polyphony} \label{section:4}

Polyphony is the condition in which a model's performance is enabled by a motley mix of mechanisms of uneven reliability, whose contributions may combine, compete, or compensate for one another. In \S\ref{section:1}, we introduced the notion through the jury room analogy. We now develop that analogy into a working model of transformer computation (\ref{subsection:4.1}) and assemble the mechanistic evidence that LLMs are polyphonic at the level of features, connections between features, and circuits (\ref{subsection:4.2}). We then show how this organisation disrupts the inference patterns surrounding the concept of understanding (\ref{subsection:4.3}), before gathering the resulting conditions into a polyphonic conception of AI understanding (\ref{subsection:4.4}).

\subsection{The Jury Room as a Model of Transformer Computation} \label{subsection:4.1}

We can now map the relay jury of \S\ref{section:1} more precisely onto a transformer. The fresh panel of jurors that comes in every morning corresponds to one layer of the network. The notes accumulated on the communal table correspond to the features active in the network’s residual stream.

Just as each juror begins by selecting material according to a certain rule or perspective, so attention heads in a transformer identify relevant information from a particular perspective, attending, for example, to numerical regularities while ignoring everything else. These attention heads often overlap in what they select.

This is followed by an information processing phase, where the jurors work in parallel to retrieve background information, flag inconsistencies, or copy salient details for later work. While some of those contributions are sophisticated, many reflect superficial heuristics or are even mistaken. But all of these contributions end up on the table for the next set of jurors to work from. This processing stage corresponds to the work of multilayer perceptrons (MLPs), which transform and enrich the information selected by the attention heads.

As the days go by, multiple lines of inquiry emerge, sometimes reinforcing one another, sometimes contradicting or interfering with one another. When a group of jurors collaborates over several days to execute what is naturally regarded as one unified procedure, this corresponds to a circuit in the network. Yet any juror can contribute to several circuits at once.

Significantly, no juror sees the complete picture. What understanding of the case emerges resides in the cumulative interplay of all these parallel contributions. No presiding juror was designated at the beginning, so where there is coordination among the contributors, it arises \textit{sua sponte}.

Yet by the final day, the jury must deliver a verdict with one voice. That verdict corresponds to the next token prediction. It may be that this verdict coincides with the one a lone expert with a comprehensive grasp of the case would have reached. But the relay jury's best work cannot be isolated from the surrounding tangle of simpler inferences and heuristics, because they are held up and enabled and shaped by them. The verdict is irreducibly a collective achievement.

Table~\ref{tab:jury} summarises the correspondences.

\begin{table}[h!]
\centering
\small
\begin{tabular}{p{0.44\linewidth}p{0.46\linewidth}}
\hline
\textbf{Jury room element} & \textbf{LLM component} \\
\hline
Communal table & Residual stream \\
A note on the table & A feature \\
Each day and panel of jurors & One layer \\
New jury each day & No memory between layers (except what is in the residual stream) \\
Selecting evidence & Attention heads \\
Processing evidence, recalling information & MLPs \\
A group of jurors performing one procedure across several days & A circuit \\
Final verdict & Model output (next predicted token) \\
\hline
\end{tabular}
\caption{Summary of the jury room analogy.}
\label{tab:jury}
\end{table}

\subsection{The Mechanistic Case for Polyphony} \label{subsection:4.2}

\begin{figure}[h]
    \centering
    \includegraphics[width=\linewidth]{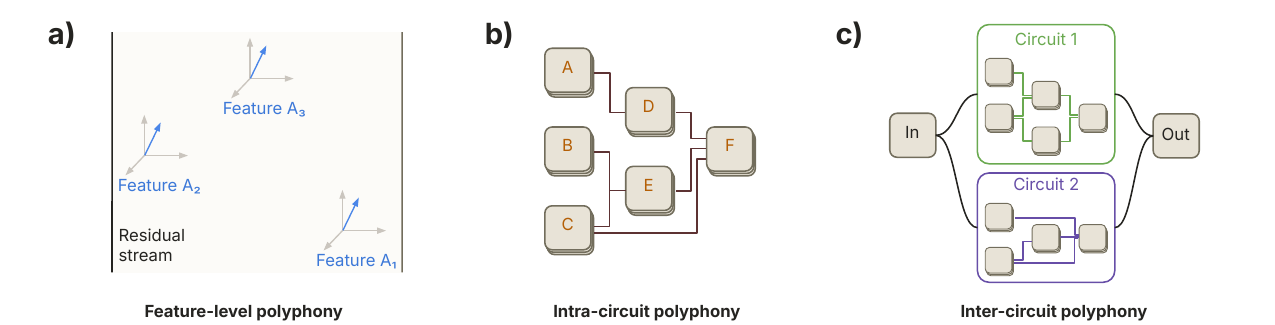}
    \caption{Polyphony at three levels: (a) \textit{features}, where related features jointly encode a concept; (b) \textit{connections}, where direct, indirect, and shortcut routes among features converge on the same output; and (c) \textit{circuits}, where two low-overlap combinations of features and connections are each sufficient to implement the procedure, and one expands its contribution when the other is ablated.}
    \label{fig:polyphony}
\end{figure}

The jury room analogy gives intuitive form to what is, at bottom, a mechanistic thesis: LLMs are polyphonic at several levels. A single content may be carried by multiple features; within a circuit, multiple pathways may make complementary, redundant, or opposing contributions; and distinct circuits may perform the same task (Fig.~\ref{fig:polyphony}).

First, there is polyphony at the level of \textit{features}. What a human would subsume under a single concept often appears in LLMs as a whole family of cognate features playing broadly the same computational role. As with the \textit{preeclampsia} feature, what a tidied-up attribution graph collects under the heading of the \textit{Texas} feature actually presents as multiple Texas-related features that \textit{collectively} perform the role of activating ``the'' Texas feature \parencite{lindsey_biology_2025}. Researchers also find features recurring at successive stages of processing and in states associated with neighbouring prompt tokens \parencite{lindsey2024crosscoders, lindsey_biology_2025}. Such recurrence may reflect real duplication, or an important signal being carried forward, or a concept being differentiated into more fine-grained ones. Talk of ``the'' feature for X is thus often shorthand for multiple representations that either redundantly, successively, or jointly encode one concept. Since this accumulation of features risks filling up the residual stream, some attention heads and MLPs act as cleanup mechanisms, cancelling selected activations by writing their negations into the stream \parencite{elhage_mathematical_2021}.

Second, there is polyphony at the level of connections, or \textit{intra-circuit polyphony}. A circuit may comprise several pathways, each consisting of a different sequence of connections between features and carrying out part of the overall computation.
When Claude 3.5 Haiku completes ``The capital of the state containing Dallas is'' with ``Austin'', for example, Dallas-related features activate Texas-related features, which combine with features directing the model to name a capital. But there is also a shortcut from Dallas to Austin; and both the Texas and ``say a capital'' clusters affect the output directly as well as indirectly via a further ``say Austin'' cluster \parencite{lindsey_biology_2025}. One feature can thus influence the output through several convergent routes.

Another example of intra-circuit polyphony is the way Claude performs two-digit addition using a coalition of heuristics \parencite{ameisen_circuit_2025, lindsey_biology_2025}. For $36 + 59$, one pathway uses stored associations between addition problems and their answers to estimate that the sum is somewhere around 92, while another considers only the final digits to infer that the answer ends in 5. Each pathway supplies a constraint, and only once these intersect (“a number near 92, ending in 5”) does the correct answer (95) emerge. Accordingly, these are not distinct circuits, but complementary pathways within a \textit{coalitional computation}.

Intra-circuit polyphony enables pathways to compensate for one another -- a phenomenon that \textcite{mcgrath2023hydra} have dubbed ``the hydra effect''. To complete ``When John and Mary went to the shop, John gave a drink to \dots'', GPT-2 Small must retrieve ``Mary''. Attention heads copy the appropriate name into the output prediction. If one head is suppressed, backup heads in later layers compensate \parencite{wang_interpretability_2023, mcdougall_copy_2024}.

Third, there is also abundant evidence of \textit{inter-circuit polyphony} -- individually sufficient circuits performing the same task. In their aptly titled ``All Circuits Lead to Rome'' paper, \textcite{chen2026allcircuits} offer the clearest challenge to the idea that LLM understanding must have one privileged internal home. They search GPT-2 for a small subgraph that can sustain performance on a task when all other connections are masked, then repeat the search while penalising the reuse of connections. Were there such a thing as \textit{the} circuit underpinning the model's performance, the searches should converge on it. Yet across twenty searches, they find multiple circuits with little overlap. Even within an apparent three-connection core, no connection is indispensable: exclude any one, and the wider model supplies another sufficient route. Each discovered circuit is therefore only one of a larger repertoire of sufficient mechanisms. This undermines the monophonic expectation that successful performance must ultimately rest on a single privileged mechanism.

Polyphony at the level of circuits also creates possibilities for \textit{destructive interference} between circuits. \textcite{kim2025reasoningcircuitslanguagemodels} and \textcite{valentino2025mitigatingcontenteffectsreasoning} identified a circuit that correctly evaluates syllogistic inferences based on logical form, but whose voice gets drowned out by the contributions of content-based mechanisms. The model is less likely to approve the syllogism: ``All apples are vegetations. All vegetations are institutions. Some apples are institutions'' than: ``All apples are edible fruits. All edible fruits are fruits. All apples are fruits''. Both are valid. But because the former contains substantively implausible premises, the content-based circuits interfere with the formal circuit's correct verdict.

Because there is polyphony both within and between circuits, a successful computation can be \textit{compositionally polyphonic}, \textit{realisationally polyphonic}, or both (Fig.~\ref{fig:failures}a). It is \textit{compositionally polyphonic} when no one contribution is sufficient and several must operate as a coalition. It is \textit{realisationally polyphonic} when it is realised by more than one circuit, so suppressing one leaves others able to take over. And it is both when it performs a task using multiple alternative coalitions that are each internally dependent on a plurality of pathways.

There is, then, plenty of mechanistic evidence of polyphony in transformer-based LLMs. Indeed, the transformer architecture may itself be conducive to polyphony. Attention heads and MLPs write additively to a shared residual stream, so their contributions remain available to downstream components unless actively negated \parencite{elhage_mathematical_2021}. Multiple partial, redundant, or conflicting signals can therefore coexist and interact without being reconciled by a central coordinating mechanism. Though not imposed by the architecture, polyphony is given room to develop in it. And one would expect the wide variety of tasks that generalist models are trained on to favour its emergence.

\subsection{How Polyphony Complicates Attributions of Understanding} \label{subsection:4.3}

The pervasiveness of polyphony means that we cannot count on identifying mechanisms that are consistently involved whenever a model performs a certain task. We may wish to identify \textit{the} mechanism, since that would give us an intellectual and practical handle on these forbiddingly complex systems. Yet even when their internal organisation becomes interpretable, it resists reduction to a single mechanism.

For the debate over AI understanding, the significance of polyphony lies in how it disrupts inferences that our inherited concept of understanding encourages us to draw -- inferences that are defeasible already in the human context, but that become downright hazardous in pervasively polyphonic systems like LLMs.

\begin{figure}[h]
    \centering
    \includegraphics[width=\linewidth]{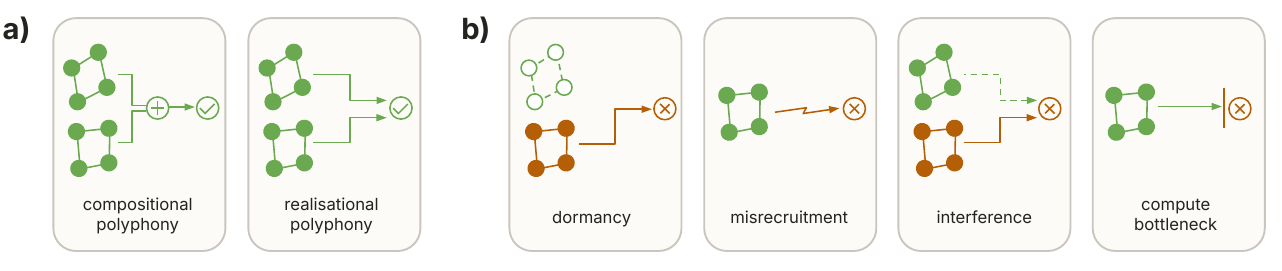}
    \caption{Sound circuitry may be compositionally polyphonic, where multiple pathways combine and none is sufficient alone, or realisationally polyphonic, where several alternative circuits can each perform the task (a). Even when sound circuitry is present, it might not bear on the output: it may be dormant, never activated; misrecruited, activated but applied incorrectly; drowned out by interference from competing mechanisms; or constrained by a compute bottleneck at inference time (b).}
    \label{fig:failures}
\end{figure}

Consider the inference from the failure to perform to the absence of understanding. There are several reasons why sound circuitry may be present without finding outward expression (Fig.~\ref{fig:failures}b).

A first possibility is \textit{dormancy}: suitable circuitry is present, but not activated. \textcite{nikankin_arithmetic_2025} find that Llama-3 relies on a collection of arithmetic heuristics that fire only over a restricted range of operands. Dormancy can also be purposely induced. Adding a suffix to a dangerous prompt can inhibit the activation of a feature that would ordinarily recruit a refusal mechanism \parencite{arditi_refusal_2024, ball2026jailbreak}.

A second possibility is what might be called \textit{misrecruitment}: not a failure to recruit sound circuitry, but a failure in recruiting it -- it is active, but not brought to bear on the case in the right way. This can happen notably when the mechanisms that bring activated circuitry to bear on the case at hand are overindexed on the training regime. Call this \textit{misrecruitment due to overfitting mediating mechanisms}. The case of ``TaxiGPT'' illustrates this. \textcite{vafa2024world} trained a GPT-2-style transformer from scratch to predict a Manhattan taxi's next turn from its origin, destination, and preceding turns (``N NW NE E SW\dots''). Although it produced legal turns nearly all the time, it occasionally implied physically impossible street configurations -- such as streets labelled NW but facing east -- or jumps over intervening streets. Performance also deteriorated sharply when the taxi was forced by researchers to turn away from its destination three quarters of the time. Vafa et al.\ concluded that the model was ``very far from recovering the true street map of New York City'' (\citeyear{vafa2024world}, 2).

Yet Beckmann, Queloz, and Freitas's (\citeyear{beckmann2026taxi}) mechanistic diagnosis of TaxiGPT's failures undercuts this inference from behavioural error to a lack of navigational understanding. They recovered a faithful map of Manhattan from TaxiGPT's activations and traced the model's errors to a problem arising downstream of forming a faithful map. On in-distribution tests, the model was able to correctly locate itself on the map. But on harder tests -- when the destination was twice as distant as during training, or on the aforementioned detour test -- the model writes its position on the map too weakly, and as noise accumulates across the map features, a wrong intersection can become active and produce an illegal move. This is \textit{misrecruitment} rather than dormancy: an accurate map is present and recruited, but the mechanism locating the taxi within it does not generalise robustly out of distribution. The problem is thus not that TaxiGPT lacks a correct map of Manhattan. The map circuitry lies ready to be recruited, but because the localisation mechanism fails on out-of-distribution cases, it is not always recruited correctly.

A third possible problem is \textit{interference}: relevant circuitry is activated, but its contribution is dominated by less reliable mechanisms. This is what happened with the circuit evaluating the formal validity of syllogisms. \textcite{rai2025parentheses} similarly find that language models can err on parentheses-balancing tasks when unreliable mechanisms swamp more reliable ones. Amplifying the reliable mechanisms can restore accuracy from approximately zero to nearly 100\%. The original failure therefore did not show that the model lacked sound circuitry. It was present and active, but outvoted.

\textcite{milliere2026anthropocentric} identify a fourth possible problem: \textit{compute bottlenecks} at inference-time. A transformer required to answer in a single forward pass is limited in what it can compute \parencite{merrill2024expressive}. A failure may therefore reflect insufficient computational room to use and coordinate the model's existing resources. Increasing a model's budget of thinking tokens can lift this constraint and remedy the failure \parencite{wei2022chain, muennighoff2025s1}.

In the terms of the jury-room analogy, the four obstacles are the following: the expert juror may be missing altogether (no sound circuitry), or present but silent (dormancy), bring their expertise to bear on a garbled characterisation of the case (misrecruitment), speak but be outvoted (interference), or lack enough time to complete and coordinate the relevant reasoning (compute bottleneck). A mistaken verdict cannot tell us which of these problems obtained and therefore does not establish that the relevant understanding was absent.

Yet the converse inference -- from successful performance to a general and reliable basis for it -- is rendered equally hazardous by polyphony. \textcite{eshuijs2025attention} showed this by training GPT-2 to classify movie reviews according to whether they expressed positive or negative sentiments. The model classified reviews reliably. Yet mechanistic analysis revealed that the model was relying on correlations in the data between actor names and sentiment: attention heads used the actor's name to steer the verdict before the review was fully processed. Replacing the actor name with one correlated with the opposite sentiment often flipped the classification. Successful model performance may thus rest on shortcuts whose unreliability emerges only once incidental correlations break.

Polyphony also weakens the correlation and integration between the different capacities conferred by understanding. With humans, there is an imperfect, but noteworthy correlation between the capacities to \textit{execute}, \textit{explain}, \textit{predict}, \textit{diagnose errors} in, and \textit{reason counterfactually} about a task one understands. A person's ability to explain the task or diagnose mistakes is therefore reasonably treated as evidence that they can apply the relevant principles themselves. Moreover, the principles they invoke in explaining the task normally also guide their own performance. The relation among these capacities is thus not merely an accidental correlation: the capacities are intelligibly coordinated. Call this expected correlation and coordination among the capacities associated with understanding \textit{cross-capacity integration}.

These capacities should be expected to come apart more readily in pervasively polyphonic systems, since the different capacities may be supported by completely distinct and isolated circuitry. Claude's polyphonic addition strategy offers an example: asked how it arrived at 95 when adding $36 + 59$, Claude invokes the schoolbook procedure of adding the ones, carrying, and then adding the tens. That is a sound method for performing the calculation, but mechanistic analysis shows that this was not how Claude performed it. The capacities to perform and explain the calculation are both displayed, but they are not integrated.

\textcite{mancoridis2025potemkin} provide a still more striking dissociation. Asked what an ABAB rhyme scheme is, GPT-4o correctly explains that the first and third lines must rhyme, as must the second and fourth. Yet when prompted to complete an ABAB poem whose first line ends with ``out'', it ends the third line with ``soft''. Asked whether ``out'' rhymes with ``soft'', the model correctly answers that it does not. The relevant principles are therefore available for explanation and error diagnosis without acquiring sufficient control over generation. The display of one capacity associated with understanding thus provides no guarantee that the others are present.\footnote{\textcite{dehghanighobadi2025explain} provide another example.}

Mancoridis et al.\ interpret this to mean that the LLM offers only an \textit{illusion} of understanding -- \textit{Potemkin understanding}, as they call it, in reference to the fake village fronts that Russian minister Grigory Potemkin is said to have erected along the Dnieper river to impress Catherine the Great. But it may be more accurate to interpret this as an example of \textit{polyphonic} understanding. After all, there is not \textit{nothing} behind the facade: the model was able to explain the rhyme scheme and diagnose errors in its execution. The behavioural evidence available cannot tell us whether the appropriate circuitry is completely absent, or present but unrecruited, or recruited but misapplied, or correctly applied but overridden.

While these hazards arise with attributions of understanding based on behavioural evidence, some inferences based on mechanistic evidence are also disrupted by polyphony. Suppose interpretability research reveals that a model performs a task using a circuit we painstakingly reverse-engineered and showed to be sound. Suppose we then encounter some further cases in which that circuit remains idle. This makes it tempting to conclude that the model’s output in those cases does not issue from sound circuitry, and hence that it does not understand. Call this the \textit{localisation fallacy}:
\arghead{The Localisation Fallacy}
\begin{argument}
\item In a set of cases $S$, the model performs task $X$ related to domain $D$ using sound circuit $C$.
\item In a set of cases $S'$, $C$ is not recruited to perform X.
\item Therefore, in $S'$, no sound circuitry underlies the model's performance of $X$.
\item Therefore, the model does not understand $D$ in the cases in $S'$, which casts doubt on its understanding of $D$ altogether.
\end{argument}
\noindent The inference from (1) and (2) to (3) is valid only given the suppressed premise that if the model’s output on $X$ issues from sound circuitry, that circuitry is $C$. But this is effectively an assumption of monophony -- a demand for a single locus of understanding. In polyphonic systems, we must be mindful of the fact that sufficiency does not entail indispensability. Understanding can be dispersed and multiply realised across the network. Localising that understanding to \textit{a} circuit need not mean that it is \textit{the} circuit. Simply monitoring the model’s use of that circuit therefore will not do. Even when that circuit remains unused or ineffective, different parts of the network may be at work. \textit{Realisational} polyphony is thus what fundamentally renders the localisation fallacy fallacious.

But the other form of polyphony we described above, \textit{compositional} polyphony, gives rise to a complementary fallacy: when we discover that an agent's performance involves a cheap heuristic, we are often quick to conclude that they do not really possess the relevant understanding. With human beings, this is often a reliable way of reasoning. If the teacher discovers that the pupil solves arithmetic problems described in vignettes by following superficial verbal cues (adding whenever he sees ``more'' and subtracting whenever he sees ``less'', say), this gives the teacher good reason to doubt that the pupil truly understands the quantitative relationships described.

To extend this inference pattern to LLMs would be to conclude, upon finding that they rely on a cheap heuristic, that their answers do not express understanding. But the difficulty with pervasively polyphonic systems is that, unbeknownst to us, the heuristic may coordinate its interaction with other pathways in such a way as to turn individual imprecision into joint precision. There is therefore a risk of treating heuristics as \textit{alternatives} to sound circuitry when they are \textit{constituents} of it. Call the premature inference from the involvement of heuristics to the absence of understanding the \textit{heuristic exclusion fallacy}:

\arghead{The Heuristic Exclusion Fallacy}
\begin{argument}
\item In a set of cases $S$, the model's performance of task $X$ related to domain $D$ causally depends in part on a cheap heuristic $H$.
\item Taken by itself, $H$ would not constitute sound circuitry for $X$ across $S$.
\item Therefore, these performances do not issue from sound circuitry.
\item Therefore, these performances do not manifest understanding of $D$ with respect to $X$.
\end{argument}

\noindent This reasoning pattern treats the discovery of a contributing heuristic as evidence of the absence of sound circuitry. Yet, as Claude's addition strategy in \S\ref{subsection:4.2} illustrated, compositional polyphony undercuts this assumption of exclusiveness. A heuristic can form part of a coalition that amounts to sound circuitry because other pathways supply complementary constraints.

The way we think about understanding in LLMs accordingly needs to be de-monophonised not just at the level of the conclusions we draw from behaviour, but all the way down to those we draw from mechanistic findings.

\subsection{A Polyphonic Conception of AI Understanding} \label{subsection:4.4}

Polyphony does not make attributions of understanding impossible, but it complicates them. A conception of understanding fitted to polyphonic systems must be sensitive to these complications. What that conception needs to do for us, after all, is to help practitioners distinguish trustworthy from untrustworthy outputs. Only a \textit{modal} conception of understanding can accomplish this, since it must be attuned not just to how the model did on past cases, but to the organisation that persists inside the model and thus makes it likely to succeed in the case at hand.

That internal organisation amounts to sound circuitry when it combines features tracking elements of a domain, connections encoding relevant dependencies between them, and circuits implementing procedures for exploiting that structure. We can derive four desiderata on such sound circuitry from the complications introduced by polyphony.

The first and most obvious condition is that sound circuitry must be \textit{present}. The model must contain some combination of features, connections, and circuits that preserves, to the degree required by the task, the distinctions and dependencies on which good performance hinges. It may well be that the model contains multiple bits of circuitry that meet this desideratum. What the presence condition requires is only that at least one instance of sound circuitry be available somewhere in the neural network. There need not be a particular region of the network that is involved in every successful performance of this kind.

Second, sound circuitry must be \textit{recruited}. It must be activated by the case at hand and not just lie dormant. As we saw, models sometimes form sound circuitry that is reliably engaged by cases within a certain range, but not by those outside that range.

Third, as the TaxiGPT example drives home, sound circuitry must be recruited \textit{correctly}. If the mechanisms mediating its application to a novel case are insufficiently reliable, this vitiates the output, even if sound circuitry is present and recruited. As we put it, sound circuitry can be \textit{misrecruited}.

Fourth, the correctly recruited circuitry must be \textit{in control} of the output. That is to say, its causal contribution must carry all the way through to what the model eventually says or does. This need not take the monophonic form of a single mechanism prevailing every time – the governing circuitry can be a coalition of mechanisms whose contributions are jointly sufficient, although no contribution is sufficient on its own. What the control condition excludes are cases in which the contributions of sound circuitry are outweighed by unsound competitors.

These nuances enable us to distinguish the \textit{depth} of a model's understanding from the \textit{hold} that this understanding has over the model's behaviour. Depth of understanding grows with the accuracy and generality of the circuitry. Its hold concerns the reliability with which that circuitry is correctly recruited and determinative of model behaviour.

Behavioural evidence alone struggles to tell the two apart. Shallow understanding with a firm hold will typically manifest as patchy performance. But the same is true of deep understanding with a precarious hold. Only a mechanistic inspection of the model can factor a performance into the components that produced it.

An attribution of understanding is inherently projective. It does not merely certify that one output issued from sound circuitry; it claims that outputs can be relied upon to do so across some range of cases and conditions of use. Call this range the attribution's \textit{scope}. In polyphonic systems, scope is best regarded as ranging only over cases and conditions of use, and not over tasks or capacities. Explaining a procedure and executing it are importantly different capacities, and whether understanding manifested in one capacity carries over to another is a further empirical question -- a question of what we called cross-capacity integration.

Among the conditions of use that delimit scope, one of the more significant is the model's inference regime. When a model is forced to respond immediately, without using many intermediate reasoning tokens, it has little room to recruit and coordinate what circuitry it harbours. Given a more generous budget of intermediate tokens to work with, however, the same model can make its circuitry go further \parencite{wei2022chain, muennighoff2025s1}. Scope is thus not determined solely by a model's weights. Equipping a model with a scratchpad for intermediate reasoning (and the possibilities for tentative reasoning, backtracking, and self-correction it provides) can significantly widen the range of cases over which the model's existing circuitry acquires behavioural hold.

Failures that look indistinguishable at the level of behaviour thus differentiate into different problems calling for different remedies. Absent or unsound circuitry calls for deeper pre-training. Dormant, misrecruited, or outvoted circuitry calls for better orchestration of the sound circuitry that has already formed, for which post-training may suffice. Insufficient computational room calls for a more generous inference regime.

In light of the above, we are now in a position to articulate a conception of understanding that fits polyphonic AI systems like LLMs:

\begin{quote}
\begin{tcolorbox}[defbox]
\textbf{AI Understanding (Def.):} A model \textit{understands} what it is doing in a given case when its output issues from sound circuitry for the task at hand -- some constellation of features, connections, and circuits that this case has correctly recruited and that controls what the model says or does. That constellation may be a coalition whose contributions are jointly sufficient without any one being sufficient alone, and it need not be the only such constellation the model harbours. More generally, a model understands a domain (with respect to a task or capacity and across a specified scope of cases and inference conditions) insofar as its outputs issue from sound circuitry in this way, though not necessarily from the same circuitry in every case. The understanding so attributed has a \textit{depth}, determined by the accuracy and generality of the circuitry that issues the output, and a \textit{hold}, determined by how reliably some such circuitry is correctly recruited and in control across that scope. Such an attribution provides defeasible, \textit{pro tanto} warrant for trusting the model's outputs within that scope.
\end{tcolorbox}
\end{quote}
This conception of understanding preserves the inferential connection that \S\ref{section:2} cast as indispensable: attributions of understanding guide the allocation of epistemic trust. It also gives attributions of understanding an empirical foothold in the internal organisation of LLMs. What the conception relinquishes is the presumption that understanding must have one privileged internal home and manifest itself uniformly across capacities. The definition acknowledges not only that the circuitry from which an output issues may be \textit{compositionally} polyphonic -- a coalition of pathways that only jointly become sufficient -- but also that the model's understanding may itself be \textit{realisationally} polyphonic, variously discharged by a plurality of different and perhaps partly overlapping coalitions. What ultimately warrants trust is not that the model always recruit the same mechanism, but that it possess a persistent capacity to recruit \textit{some} sound circuitry correctly and to give that circuitry sufficient control over its behaviour.

Far from closing off the question of AI understanding, then, the phenomenon of polyphony transforms it into a series of questions that can be addressed through mechanistic investigation -- at least in principle. In the next section, we turn to what this means in practice, when the question arises at the scale of frontier models.

\section{Assessing Understanding in Frontier Models} \label{section:5}

The evidence we considered came from relatively small models, and one might object that this tells us little about frontier models, which are vastly larger. This size difference suggests two reasons for pessimism: large models have more space to memorise what they are trained on, and they are too complex to reverse-engineer in full.

While the difficulty is real, it does not render our proposed conception of AI understanding idle. Evidence for the presence, recruitment, and control of sound circuitry admits of degrees, and it can be assembled even here.

The findings from smaller models have a limited, but important role in this. For a given task and domain, they establish that transformer architectures \textit{can} form features tracking domain structure, connect them in appropriate ways, and combine them into procedures that generalise beyond memorised examples. These findings act as possibility proofs. If smaller models can do it, larger models ought to be able to do it, too.

In fact, evidence suggests that larger models tend, if anything, to generalise even better than smaller ones. The ``lottery ticket hypothesis'' surmised that since a bigger network contains more sub-networks, it is likelier to harbour a winner \parencite{frankle2019lottery}. Yet newer research offers more systematic explanations. One is that larger models are less likely to get trapped in a local minimum, because the added width also widens traps in the loss landscape, opening escape routes towards better solutions \parencite{simsek2021geometry, martinelli2026landscape}. Another is that larger models are more likely to land on simpler or more compressible solutions, because these take up a larger share of the space \parencite{wilson2025deep}.\footnote{Among the weight-settings that fit the training data, some sit on narrow spikes, and others in broad basins, where a whole neighbourhood of settings does equally well. The basins constitute more compressible solutions, since there is no need to specify them precisely; and since volume compounds across dimensions, every parameter added increases the compressible solutions' share of the space.} A third explanation is that the extra capacity lets models form circuitry for rare tasks, which are harder to learn because they provide weaker training signals, and which a smaller model has to neglect in favour of frequent tasks \parencite{huang2026rare}.

These considerations weaken the first reason for pessimism: scale need not favour memorisation over generalisation. But being better equipped to form general procedures is not the same as actually forming and appropriately recruiting, and relying on them. The second reason for pessimism is therefore more pressing.

Yet full reconstruction is not necessary for every attribution of understanding. It may often be enough to investigate how a model performs a specific type of task. In particular, what is needed is enough evidence to assess whether the conditions specificed in \S\ref{section:4} are met:

\begin{enumerate}
\item \textit{Presence.} Does the model contain sound circuitry for the task at hand -- some constellation of features, connections, and circuits that this case has correctly recruited and that controls what the model says or does? One indicator of this would be that the relevant domain structure can be decoded from the model's internal states. Because of polyphony, that circuitry can be a coalition of jointly sufficient but not individually necessary mechanisms, and there may be several such coalitions. 
\item \textit{Recruitment.} Is that sound circuitry reliably activated, or does it remain dormant outside a certain range of inputs?
\item \textit{Control.} When sound circuitry is recruited, is it recruited correctly, and does its contribution govern the output, or does it get outweighed by competing mechanisms?
\item \textit{Scope.} For what range of cases do the presence, recruitment, and control conditions continue to obtain? Does the model's competence extend across the capacities for execution, explanation, prediction, error diagnosis, and counterfactual reasoning?
\end{enumerate}

Ideally, a model's fulfilment of these conditions would be evaluated using mechanistic evidence. But even in the absence of mechanistic interpretability tools, behavioural indicators offer some hint, if not conclusive evidence, of the extent to which a model meets these conditions: one can test the model on challengingly novel cases, check whether changing the framing in ways that should be immaterial affect the model's performance, and see whether it is appropriately sensitive to changes in framing that are relevant and that should change the model's response. Moreover, testing the model's ability to integrate its understanding across different capacities reveals the degree of its cross-capacity integration.

In addition, the system cards that many leading labs publish with each new model release increasingly include mechanistic evidence to show that relevant features and circuits were identified and shown to predictably alter the model's behaviour. This points to a future where system cards might publicly certify and advertise a model's governance by sound circuitry for certain tasks and domains. We are used to testing human understanding through exams and collections of hard problems. Yet the study of human understanding has not had the benefit of the tools and possibilities that interpretability research now has at its disposal. With the repertoire of white-box methods growing at pace, we should expect the study of AI understanding to push far beyond behavioural evaluations, in ways that human-facing attributions of understanding never could.

\section{The Orchestration Gap} \label{section:6}

Whether AI models understand, we have argued, is a question we ultimately cannot avoid confronting in practice. The conceptual neeed to distinguish trustworthy from untrustworthy AI outputs cannot be met by purely mathematical or statistical descriptions. And, as our reintroduction argument indicated, any vocabulary rich enough to meet this need reforges at least some of the inferential connections that charactise the concept of understanding. Yet we are hindered in this by the tendency to envision understanding as something unified and localisable to a particular area underpinning execution, explanation, prediction, diagnosis, and counterfactual reasoning. This may be a tolerable idealisation of human understanding, but it becomes a real obstacle when dealing with systems as polyphonic as current LLMs.

The positive proposal of this paper is a conception of AI understanding centred on sound circuitry. A model understands something when it contains some constellation of features, connections, and circuits that bears a non-accidental, approximately structure-preserving relation to the task-relevant organisation of the domain, and that would meet an independently specified standard if correctly recruited and allowed to govern the computation. Any such attribution must therefore specify a task and scope; within that scope, the relevant circuitry must be present, reliably and correctly recruited, and in control. A model's depth of understanding concerns the richness, accuracy, and generality of its sound circuitry; its hold concerns how reliably that circuitry is recruited and governs what the model says or does. Because sound circuitry may consist of a coalition of individually insufficient components, and because several alternative coalitions may perform the same task, understanding need have no privileged internal home.

Polyphony is not peculiar to LLMs.\footnote{Neuroscience even has a name for a comparable phenomenon in the human brain: ``degeneracy'', the capacity of structurally different elements to perform the same function \parencite{edelman2001degeneracy}. Degeneracy allows distinct neural pathways to sustain the same cognitive capacity and confers robustness when one route fails \parencite{price2002degeneracy, noppeney2004degeneracy}. This is related to \textit{graceful degradation}, which LLMs also exhibit: sizeable blocks of layers can be ablated with surprisingly modest loss \parencite{michel2019sixteen, gromov2024unreasonable}.} People confident that they understand a familiar mechanism quickly discover how fragmentary their understanding really is when asked to explain it \parencite{rozenblit2002illusion}. But we have various techniques for orchestrating these pieces and half-remembered schemas into something more unified. What remains of Kepler's laws years after a class may be disconnected fragments, resembling what \textcite{freeborn2026fractured} calls ``fractured understanding'' in deep nets. When prompted to explain those laws to one's children and use them to predict a planet's position, however, we reach for pen and paper, testing and revising until the pieces cohere. Nor is this peculiar to the layperson: \textcite{boge2026understanding} argue that even a scientist's understanding of a phenomenon is often only a tacit model, and counts as properly \textit{scientific} understanding only once it coalesces into an explanation that the scientific community can follow. On this picture, the orchestration of polyphony is not so much a standing feature of our cognitive architecture, but a continual achievement -- the successful coordination of partial resources, often with social and material scaffolding.

There is emerging evidence that scaffolding can similarly improve orchestration in LLMs. External scratchpads and agentic harnesses provide a workspace in which different parts of a model can be successively recruited, tested, and recombined. \textcite{venhoff2026reasoning} offer a striking example. They found that the chain-of-thought of a fully trained reasoning model decomposes into a small repertoire of distinct moves, such as planning, checking, and backtracking. Interestingly, those activation signatures were already present in the base model. A simple classifier was then trained to select which move should come next and nudge the base model's activations accordingly; this intervention closed most of the gap between the two models. What the base model lacked was not the capacity to make the individual moves, but reliable control over when to deploy them. This is the \textit{orchestration gap}: a failure to coordinate competences that the system already possesses.

Current LLMs nevertheless do not orchestrate their partial competences as reliably as humans.\footnote{As illustrated by the alien failure modes LLMs continue to display \citep[e.g.,][]{mccoy2024embers, chen2025alien}.} Until this gap closes, practitioners cannot rely on the familiar inferences associated with the monophonic conception of understanding. But they can rely on those licensed by a polyphonic conception, which turns the elusive question of AI understanding into tractable empirical questions about the presence, recruitment, and control of sound circuitry -- questions open to mechanistic investigation and intervention. If the relay jury's verdict conceals the polyphony that produced it, we must look to the notes on the table to tell whether that verdict deserves our trust.

\clearpage

\bibliographystyle{apalike}
\bibliography{refs}

\end{document}